\documentclass[9pt,twocolumn,twoside]{osajnl}

\journal{josaa} 

\setboolean{shortarticle}{false}

\ifthenelse{\boolean{shortarticle}}{\colorlet{color2}{color2b}}{\colorlet{color2}{color2}} 

\title{Interactive Removal and Ground Truth for Difficult Shadow Scenes}

\author[1,*]{Han Gong}
\author[2]{Darren Cosker}

\affil[1]{School of Computing Sciences, University of East Anglia, UK}
\affil[2]{Department of Computer Science, University of Bath, UK}

\affil[*]{Corresponding author: h.gong@uea.ac.uk}

\dates{Compiled \today}

\ociscodes{(100.0100) Image processing; (150.2950) Illumination}

\doi{\url{http://dx.doi.org/10.1364/ao.XX.XXXXXX}}

\usepackage{makecell}
\usepackage{tikz,psfrag}
\usepackage{multirow}
\usepackage{subfigure}
\usepackage[algo2e,ruled]{algorithm2e}
\usepackage{amsmath,amsfonts}
\usepackage{dsfont}
\usepackage{mathtools}
\usepackage[T1]{fontenc}
\usepackage{bookmark}
\DeclarePairedDelimiter{\ceil}{\lceil}{\rceil}

\DeclareGraphicsExtensions{.pdf}
\tikzstyle{block} = [rectangle,draw,text centered, rounded corners,above right]
\providecommand{\abs}[1]{\lvert#1\rvert}
\providecommand{\tbpic}[1]{\begin{minipage}{.13\textwidth}
\includegraphics[width=\linewidth]{#1}\end{minipage}}
\providecommand{\tbpicx}[1]{\begin{minipage}{.16\textwidth}
\includegraphics[width=\linewidth]{#1}\end{minipage}}
\providecommand{\tbbpic}[1]{\begin{minipage}{.18\textwidth}
\includegraphics[width=\linewidth]{#1}\end{minipage}}

\renewcommand\thesubsection{\thesection.\Alph{subsection}}

\usepackage{xspace}
\newcommand*{\eg}{\emph{e.g.}\@\xspace}
\newcommand*{\ie}{\emph{i.e.}\@\xspace}
\newcommand*{\etal}{\emph{et al.}\@\xspace}

\makeatletter
\newcommand*{\etc}{%
    \@ifnextchar{.}%
        {etc}%
        {etc.\@\xspace}%
}
\makeatother

\begin{abstract}
A user-centric method for fast, interactive, robust and high-quality shadow removal is presented. Our algorithm can perform detection and removal in a range of difficult cases: such as highly textured and colored shadows. To perform detection an on-the-fly learning approach is adopted guided by two rough user inputs for the pixels of the shadow and the lit area. After detection, shadow removal is performed by registering the penumbra to a normalized frame which allows us efficient estimation of non-uniform shadow illumination changes, resulting in accurate and robust removal. Another major contribution of this work is the first validated and multi-scene category ground truth for shadow removal algorithms. This data set containing 186 images eliminates inconsistencies between shadow and shadow-free images and provides a range of different shadow types such as soft, textured, colored and broken shadow. Using this data, the most thorough comparison of state-of-the-art shadow removal methods to date is performed, showing our proposed new algorithm to outperform the state-of-the-art across several measures and shadow category. To complement our dataset, an online shadow removal benchmark website is also presented to encourage future open comparisons in this challenging field of research.
\end{abstract}

\setboolean{displaycopyright}{true}

\begin{document}
\maketitle

\thispagestyle{fancy}

\ifthenelse{\boolean{shortarticle}}{\ifthenelse{\boolean{singlecolumn}}{\abscontentformatted}{\abscontent}}{}
\section{Introduction}
\label{sec:intro}
Shadows are ubiquitous in image and video data, and their removal is of interest in both computer vision and graphics. Although shadows can be useful cues, \eg shape from shading, they can also affect the performance of algorithms (\eg in segmentation and tracking). Their removal and editing is also often the pain-staking task of graphical artists. A successful shadow removal method should seamlessly relight the shadow area while keeping the lit area unchanged. The umbra is the darkest part of the shadow whilst the penumbra is the transitional shadow boundary with a non-linear intensity change between the umbra and lit area. The textures in shadowed surface generally become weaker that contrast artifacts can appear in shadow areas due to image post-processing~\cite{Arbel2011}.

One of the difficulties in detecting and removing shadows is the large variability in different shadow types. In particular, the following common attributes (\eg Fig.~\ref{fig:datashow}) of shadows can significantly increase the difficulty of their removal:
\begin{itemize}
	\item \textbf{Texture of cast surface} Strong texture causes higher intensity variation which makes it difficult to extract illumination change from intensity changes. In addition, dark textures can appear similar to shadows, which can often confuse shadow detection algorithms.
	\item \textbf{Shadow softness} Softness generally relates to the size of a shadows penumbra. Higher softness brings challenges in preserving penumbra texture when removing shadow. When the illumination change becomes much weaker than the intensity change caused by texture, it can be difficult to extract the component of illumination change.
	\item \textbf{Broken shadow} Broken shadows contain variable illumination attributes such as irregular shape, highly varying penumbra size, and overlapping penumbra. Fixed illumination models can find such irregular illumination changes difficult to process.
	\item \textbf{Shadow color} When shadows are not conventionally black but instead colorful, it is not only difficult for machines to detect this appearance change but even humans. Also, even when shadows of this kind are detected, their removal is still difficult as the color in the umbra could be related to the surfaces' reflection.
\end{itemize}

Given sufficient training data, automatic approaches can robustly remove common shadows, however there are difficulties in dealing with difficult shadow scenes. User-aided approaches give users more control over difficult shadow removal. In this paper, an interactive, high-quality and robust method for fast shadow removal is proposed using two rough user-defined strokes indicating the shadow and lit image areas. The approach sacrifices full autonomy~\cite{Guo2012,khan2016automatic} for very broad and simple user input -- contrasting with existing manual approaches that require fine-scale input (accurate shadow contours~\cite{Su2010,Liu2008}) or highly simplistic inputs (single pixel~\cite{ShorL2008}) that can result in shadow detection artifacts. This on-the-fly learning approach is robust to large variations in user input, and can remove shadows with difficult attributes such as colored and broken shadow. Given detection, reliable shadow removal is delivered -- verified with thorough quantitative tests for different types of shadow (for the first time in this area) comparing to previous state of the art approaches. A large high-quality and multi-scene-category ground-truth data set for the evaluation of shadow removal is also presented -- consisting of 186 images with reliable ground truth. This overcomes issues with previous data sets -- such as inconsistencies between shadow and shadow-free images -- the results of which are also quantitatively verified and compared against previous data sets. The approach presented represents what the authors believe to be a state-of-the-art method for shadow removal, with the most robust evaluation of such methods to date across a range of difficult shadow cases.
\subsection{Related work}
A shadow is generally defined as having an umbra and penumbra area -- denoted by the central shadow region and its border (penumbra) transitioning illumination between the fully dark and lit area.
A shadow image $I_{c}^{+}$ can be considered to be a Hadamard product~\cite{barrow1978recovering} of a shadow scale layer $\mathcal{S}_{c}$ and a shadow-free image $I_{c}^{*}$ as follows:
\begin{equation}
    \label{equ:image_formation}
    I_{c}^{+} = I_{c}^{*}\circ\mathcal{S}_{c}
\end{equation}
where $c$ is a RGB channel.
For a lit pixel, the illumination is constant in both shadow and shadow-free images. For a shadow pixel, its intensity in a shadow image is lower than its intensity in the shadow-free image. Therefore, the scales $\mathcal{S}_{c}$ of the lit area are $1$ and other areas' scales are between $0$ and $1$.

Approaches to shadow removal can be categorized as either automated or user-aided. The differentiation between fully automated or user-aided relates to initial detection of the shadow -- with removal itself (after detection) being a largely automatic task. In any case, both removal and detection are ill-posed problems and difficult to reliably achieve. We summarize the features and requirements of the recent shadow removal methods in Table~\ref{tbl:cmp}.

\begin{table}[htb!]
\small
\centering
\caption{Feature comparison of shadow removal methods: ``Illumination Preserving'' refers to the ability to preserve the original illumination in the lit area. ``Texture Preserving'' refers to the preservation of the correct surface texture under the penumbra after removal. ``Color Correction'' refers to the ability to correct color artifacts caused by image post-processing after removal.}
\label{tbl:cmp}
\input{cp_review.clo}
\end{table}

\subsubsection{Shadow detection}
As for automated shadow detection, intrinsic image~\cite{Yang2012} based methods and illumination-invariant image~\cite{Finlayson2009,fredembach2006simple,fredembach2005hamiltonian} based methods are one such popular approach to the problem. The decomposition of intrinsic images provides shading and reflectance information but can be unreliable leading to over-processed results. Intrinsic image based methods generally assume that the illumination change leads to smooth intensity change and the neighboring pixels have similar chromaticities. Illumination-invariant image is fast to compute but it only provides reflectance information. The derivation of illumination-invariant image assumes that the image is linear (not rendered). However, most of the images found on the Internet are rendered, \eg, compressed and gamma corrected. The non-linearity caused by image rendering can break the algorithm. Besides, their shadow detection relies on comparing the difference between the edge detection results of a shadow image and its illumination-invariant image. This property makes these methods incapable of removing soft or light shadows. Shadow detection can also be achieved by shadow features learning~\cite{Huang2009,Zhu2010,Lalonde2010,Guo2012,vicente2013single,huang2011characterizes,shen2015shadow,khan2016automatic}. However, shadow detection is constrained by the range of training data and quality of classifier, image edge detection and segmentation there-in. \cite{Zhu2010,Lalonde2010,Huang2009,vicente2013single,huang2011characterizes} detect shadows by classifying edges in an image according to shadow edge features such as changes in intensity, texture, and color ratio. Guo \etal \cite{Guo2012} adopt similar features but detect shadows by classifying segments in an image and pairing shadow and lit segments globally. This method is more robust because segment pairing correlates both neighboring and non-neighboring surfaces. Some recent methods~\cite{shen2015shadow,khan2016automatic} adopted Convolution Neural Network (CNN) for detecting shadows from single images. Based on training from massive data, CNN-based shadow detection provides fast speed and high accuracy. However, the science behind a CNN remains unexplainable. Some methods require additional controllable light sources to capture shadow-less objects, \eg, by comparing flash and no-flash image pairs~\cite{Drew2006}. However, active lighting restricts the applicable type of scenes - as moving lights around and using special lighting setups outdoors is often not practical. Some methods adopt optical filters to obtain multi-spectral images for illumination detection, \eg by comparing NIR and RGB images~\cite{Salamati2011} and by comparing RGB and single-color-filtered image~\cite{Finlayson2007}, but these methods assume some special scenarios, \eg sunlight and non-black surfaces. They are thus not applicable to the removal of normal single RGB images.

User-aided methods generally achieve higher accuracy in shadow detection at the practical expense of varying degrees of manual assistance. Wu \etal~\cite{Wu2007} require extensive user input where the user needs to define multiple regions of shadow, lit area, uncertainty and exclusion. They estimate the probability that a pixel is part of a shadow according to a 3D Gaussian Mixture Model (GMM)~\cite{mitchell1997machine} generated from the supplied pixel samples. Liu and Gleicher~\cite{Liu2008} require fine input defining the accurate shadow boundary. Su and Chen~\cite{Su2010} require a rougher shadow boundary as input and align the shadow boundary according to intensity gradient. However, it is cumbersome to define the boundaries of broken shadows in \cite{Liu2008} and \cite{Su2010}. Arbel and Hel-Or~\cite{Arbel2011} require users to specify multiple texture anchor points to detect a shadow mask but the number of user input can significantly increase when shadow regions are multiple and scattered. Xiao \etal~\cite{xiao2013fast} and Zhang \etal~\cite{zhang2015shadow} require two types of user scribbles for sampling shadow and lit intensities. Given enough scribbles, their detection methods can robustly produce shadow masks. However, their methods also require a shadow matte (guided by the scribbles) to identify shadows, which is sensitive to user-scribbles because their image matting is affected by pixel location. Shorl and Lischinski~\cite{ShorL2008} only require one shadow pixel as input. The algorithm detects shadow using image matting from a grown shadow seed. But it has limitations in cases where the other shadowed surfaces are not surrounded by the initially detected surface or when the penumbra is too wide.

\subsubsection{Shadow relighting}
Shadow relighting is another difficult problem, especially for the recovery of penumbra. Some methods apply zero-penumbra-gradient-filling~\cite{Finlayson2002} or native in-painting~\cite{ShorL2008,fredembach2005hamiltonian,fredembach2006simple} for penumbra recovery which result in penumbra texture loss. Finlayson \etal~\cite{finlayson2006removal,Finlayson2009} apply an iterative diffusion process to smoothly fill in the derivatives in penumbra which but can still cause texture loss in penumbra.
Liu and Gleicher~\cite{Liu2008} apply a curve fitting method and a global alignment of gradients to acquire shadow scales but has issues when relighting the umbra and can introduce artifacts at uneven boundaries. Arbel and Hel-Or~\citep{Arbel2011} apply a thin-plate model to fit the intensity surface and the algorithm is designed for removing shadows from curved surfaces. Guo \etal~\cite{Guo2012} remove shadows by image matting which treats shadows as the foreground. Their approach can usually introduce penumbra artifacts. Gong \etal~\cite{Gong2013} also apply a curve fitting model and they adopt an intelligent sampling scheme to improve the quality of intensity samples for illumination estimation. Su and Chen~\cite{Su2010} estimate shadow scales by using dynamic programming. These data fitting based methods~\cite{Liu2008,Arbel2011,Gong2013} assume highly-constrained curve or surface functions for illumination change which limit their range of removable shadows. Xiao \etal~\cite{xiao2013fast} apply a multi-scale adaptive illumination transfer which performs well for removing shadows cast on strong texture surfaces. Zhang \etal~\cite{zhang2015shadow} remove shadows by aligning the texture and illumination details between corresponding shadow and lit patches. However, both \cite{xiao2013fast} and \cite{zhang2015shadow} are very sensitive to variable user inputs. Khan \etal~\cite{khan2016automatic} apply a Bayesian formulation to robustly remove common shadows. However, this method is unable to process difficult shadows such as non-uniform (or broken) shadows. It is also computationally expensive due to a large number of unknown parameters.

\subsubsection{Shadow removal ground truth}
To date, most shadow removal methods (\eg \cite{Wu2007,Su2010,Arbel2011,Gong2013,xiao2013fast,zhang2015shadow}) have been evaluated by visual inspection on some selected images -- with only a few exceptions performing quantitative evaluation. This is in part due to a lack of high-quality, varied, and public ground truth data. Shorl and Lischinski\cite{ShorL2008} perform a quantitative test but comparison is difficult due to the their data not being publicly available. Guo \etal~\cite{Guo2012} provide the first public ground truth data set for shadow removal and perform quantitative testing. However, the difficulty of collecting such a data set is highlighted in their work, with the appearance of some global illumination changes and mis-registration between the shadow and shadow-free images being a difficult factor to control. This can make quantitative testing on such data somewhat difficult, as these errors can influence shadow removal results. 

Another desirable property as yet not explored by existing data sets or fully explored in work on detection and removal is the categorization of shadows. Such attributes are important to consider as these different shadow types can present their own unique challenges, \eg removal of colored shadows (\ie through a glass bottle) are more difficult than consistent unbroken shadows (\ie a human silhouette). Universal shadow removal approaches should therefore be capable of handling these multiple cases. In addition, having such categories in a ground truth data set is also important -- if only to allow us to evaluate different algorithm performance in a range of scenarios and scene types.

\subsection{Contributions}
Given our review of previous work, 4 main contributions are proposed:\newline
\textbf{1) A rigorous, highly-varied and categorized shadow removal ground truth data set:} Our quantitatively verified high quality data set contains 186 ground truth images organized into common shadow categories. Based on this data, our method is quantitatively evaluated against other state-of-the-art algorithms on different shadow category types.\newline
\textbf{2) Simple and robust user input based shadow detection:} Our shadow detection component requires only two rough user scribbles marking samples of lit and shadow pixels. Our approach differs from previous work requiring more complex user-inputs~\cite{Wu2007,Liu2008,Su2010,Arbel2011} or simpler inputs~\cite{ShorL2008} that compromise robustness and quality.\newline
\textbf{3) High quality and fast shadow removal:} Unlike existing methods requiring slow pixel-wise optimization~\cite{Su2010,ShorL2008,Wu2007,Guo2012,khan2016automatic} or an inflexible fitting model~\cite{Liu2008,Arbel2011,Su2010,Guo2012,zhang2015shadow,xiao2013fast,khan2016automatic}, penumbra unwrapping is introduced upon which multi-scale smoothing is performed to derive sparse shadow scales across the penumbra. This allows robust and efficient estimation of illumination changes without requiring prior training and any assumed illumination change models. This method is simple and fast yet offers state-of-the-art shadow removal quality.\newline
\textbf{4) Robust color correction:} Post-processing effects may cause inconsistency in shadow corrected areas compared with the lit areas both in tone and contrast. A robust multi-scale color correction is proposed to amend these artifacts.

To summarize, the authors believe these contributions to be important to this area of research due to their significant improvements over the state-of-the-art in shadow removal in a wide range of repeatable tests.
\section{Shadow Removal Ground Truth}
\label{sec:gt}
\begin{figure*}[!ht]
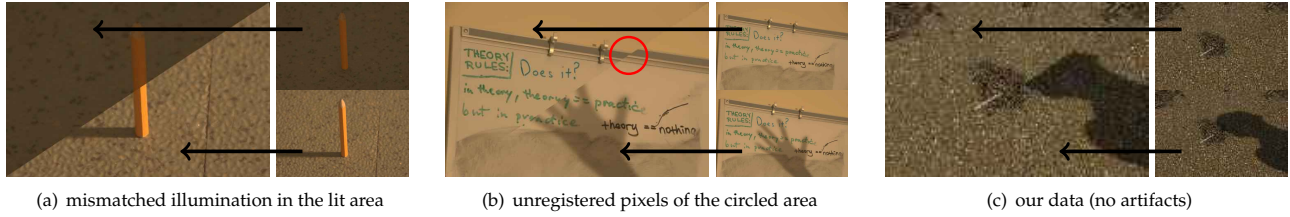

  \centering
  \vspace{-5pt}
  \subfigure[mismatched illumination in the lit area]{
    \input{gt1.clo}
    \label{fig:lit-diff}
  }
  \subfigure[unregistered pixels of the circled area]{
    \input{gt2.clo}
    \label{fig:pixel-shift}
  }
  \subfigure[our data (no artifacts)]{
    \input{gt3.clo}
    \label{fig:well-captured}
  }
  \vspace{-10pt}
  \caption{Issues of shadow-free ground truth in the previous data set~\cite{Guo2012}. To easily examine the ground truth artifacts, we extract one half from a shadow image and another half from a shadow-free ground truth image and merge these two halves. For each sub-figure: top left segment -- shadow-free image; bottom right segment -- shadow image. \subref{fig:lit-diff} and \subref{fig:pixel-shift} are taken from \cite{Guo2012} which reflect the two annotated issues. An example from our data -- which rejects image pairs with these properties is shown in \subref{fig:well-captured}.}
  \label{fig:db}
\end{figure*}

A thorough quantitative evaluation of shadow removal performance requires a high-quality, diverse shadow-free ground truth. The first public data set was supplied by \cite{Guo2012}. However, while this is a valuable resource for evaluating shadow removal and the first of its kind, there are many opportunities for expansion and several improvements are presented in our new data set. Firstly, the concept of shadow categories is introduced for the first time in our proposed data set, and a wide range of new types proposed. Secondly, ground truth is constructed and verified in a careful manner so as to remove irregularities between test and validation images. In terms of the latter, we note environmental illumination and registration errors between some shadow and ground truth images in existing data sets. An example of comparison is shown in Fig.~\ref{fig:db}. Our new data set avoids these issues and represents, we believe, the most stable and thorough data set for shadow removal evaluation available today. In order to highlight the benefits of our rigorous data protocol, in \S\ref{sec:ev} the quality of our ground truth data is quantitatively compared to another state-of-the-art dataset~\cite{Guo2012}.

Shadow images and their ground truth are captured using a camera with a tripod and a remote trigger. This rig minimizes misalignment due to camera shake. To minimize illumination variance, images are captured within a very short interval of time using a manual capture mode with fixed ISO and exposure settings. When collecting data, environmental effects are often unavoidable, \eg, wind can cause camera shake or the sun might move behind the clouds. Such failed acquisitions are rejected from our data set using a quantitative assessment outlined in \S\ref{sec:ev}. For evaluation, our shadow data is also categorized according to 4 different attributes: degree of texture, shadow softness, brokenness of shadow, and color variation. All the validated shadow images are manually categorized according to 4 shadow categories and 3 intensity degrees. The labeling was performed by 5 users and their average responses are rounded to the nearest intensity degree numbers (\eg 1 for ``weak'' and 3 for ``strong''). In total, our final data set after rejection consists of 186 test cases. For comparison, the previous state of the art from \cite{Guo2012} consists only of 28 test cases after applying our strict rejection measure. Examples of images in each category are shown in \S\ref{sec:eva-vc}.
\section{Interactive Shadow Removal}
\label{sec:ov}
In this section, our algorithm is first explained in brief. Technical details for each of its components are then expanded upon in following sections. Our algorithm consists of 4 steps (see Fig.~\ref{fig:ppl}):\newline
\textbf{1) Pre-processing (\S\ref{sec:pp})} An initial shadow mask (Fig.~\ref{fig:ppl}(b)) is detected using a KNN classifier trained from data from two rough user inputs (\eg Fig.~\ref{fig:ppl}(a)). A \emph{fusion image}, which magnifies illumination discontinuities around shadow boundaries, is generated by fusing channels of YCrCb color space and suppressing texture (Fig.~\ref{fig:ppl}(c)).\newline
\textbf{2) Penumbra Unwrapping (\S\ref{sec:pu})} Based on the detected shadow mask and fusion image, pixel intensities of sampling lines are sampled perpendicular to the shadow boundary (Fig.~\ref{fig:ppl}(d)). Noisy samples are removed and remaining columns stored as the initial penumbra strip (Fig.~\ref{fig:ppl}(e)). The initial columns' illumination changes are also aligned (Fig.~\ref{fig:ppl}(f)) by a fine-scale alignment.\newline
\textbf{3) Relighting (\S\ref{sec:ess})} From the penumbra strip, a multi-scale shadow scale estimation is applied to quickly and robustly estimate the illumination change along sampling lines and derive the sparse scales for all sampled sites (Fig.~\ref{fig:ppl}(g)) which are propagated to form a dense scale field (Fig.~\ref{fig:ppl}(h)). Shadows are removed by inverse scaling using this non-uniform field (Fig.~\ref{fig:ppl}(i)).\newline
\textbf{4) Color Correction (\S\ref{sec:gcc})} Post-processing effects may cause inconsistent tone and contrast in shadow removed areas compared with the lit areas'. Without introducing additional artifacts, a multi-scale color correction is proposed to remove these inconsistencies (Fig.~\ref{fig:ppl}(j)).
\begin{figure*}[htb!]
  \centering
  \resizebox{\linewidth}{!}{\input{dm.clo}}
  \caption{Our shadow removal pipeline. (a) input: a shadow image and user strokes (blue for lit pixels and red for shadowed pixels); (b) detected shadow mask; (c) fusion image; (d) initial penumbra sampling (the actual density of samples are higher than the displayed samples'); (e) initial penumbra unwrap (only the shadow edges of the largest shadow segment is shown); (f) further aligned penumbra unwrap; (g) sparse shadow scale; (h) dense shadow scale; (i) initial shadow removal result; (j) color corrected shadow removal result; (k) ground truth.}
  \label{fig:ppl}
\end{figure*}

Our shadow removal approach includes some standard algorithms which require parameters. These required parameters are denoted throughout the paper and are determined by genetic optimization based parameter learning in \S\ref{sec:para}. In this paper, we denote 6 undetermined parameters as $h_{1}, h_{2}, \dots, h_{6}$. 
\subsection{Pre-Processing}
\label{sec:pp}
Pre-processing provides a detected shadow mask and a fusion image to assist penumbra unwrapping. Although there have been automatic methods for shadow detection (\eg \cite{Guo2012,khan2016automatic,finlayson2002removing,Lalonde2010,Zhu2010,Huang2009}), results are dependent on training data quality and variation. Instead, our method requires no prior training or learning -- only two user-supplied rough inputs indicating sample lit and shadow pixels (Fig.~\ref{fig:ppl}(a)). Highlighted pixels' RGB intensities in the Log domain are supplied as training features and used to construct a KNN classifier (K = 3). Euclidean distance is used as the distance measure and the majority rule with nearest point tie-break as the classification measure. Spatial filtering with a Gaussian kernel (size = $h_{1}$, standard deviation = $\ceil[\big]{h_{1}/2}$) is applied to the obtained image of posterior probability and binarize the filtered image using a threshold of $0.5$ (\eg Fig.~\ref{fig:ppl}(b)).

Although detection errors along the boundary, as well as post-filtering, can result in intensity samples with unaligned illumination changes at sharp boundaries, our penumbra unwrapping and alignment step (\S\ref{sec:pu}) can compensate for this. Thus, our shadow \emph{removal} method is somewhat robust to noise in the initially detected shadow mask, and would also be applicable to alternative (\eg automatic) detection methods. While our user input format is identical to the two types of scribbles adopted in \cite{zhang2015shadow,xiao2013fast}, our method is found more robust to rougher (or fewer) user inputs since our method does require an image matting process that is also guided by sampled pixel location (see \S\ref{sec:variable_input} for a test example).

To assist unwrapping of the penumbra, an image is derived that magnifies illumination discontinuities around the shadow boundary -- also assisting penumbra location -- which is called the \emph{fusion image} (\eg Fig.~\ref{fig:ppl}(c)). There are 2 steps in this process:\newline
\textbf{1) Magnification of Illumination Discontinuity} An initial fusion image $\mathcal{F}$ is derived that maximizes the contrast between shadow and lit areas by linearly fusing the three channels ($\mathcal{C}_{l}$) of YCbCr space as follows:
\begin{equation}
    \label{equ:blending}
    \mathcal{F} = \sum\nolimits_{l=1}^{3} a_{l}\mathcal{C}_{l} \mbox{ subject to } \sum\nolimits_{l=1}^{3} a_{l} = 1
\end{equation}
where $a_{l}$ is the positive fusing factor of $\mathcal{C}_{l}$. The best fusing factors are derived by minimizing the following objective function $E_{b}$:
\begin{equation}
  \label{equ:blending_opt}
  E_{b}(\mathbf{a}) = \mu(\mathcal{F}_{\mathbf{S}})/\mu(\mathcal{F}_{\mathbf{L}}) + (\sigma(\mathcal{F}_{\mathbf{S}})+\sigma(\mathcal{F}_{\mathbf{L}}))/\sigma(\mathcal{F}_{\mathbf{S}\cup\mathbf{L}})
\end{equation}
where $\mathbf{a}$ is the vector of fusing factors and $\mathcal{F}_{\mathbf{S}}$ and $\mathcal{F}_{\mathbf{L}}$ are the two sets of shadow and lit pixels marked by user scribbles. In this paper, $\sigma$ and $\mu$ are defined as functions that respectively compute the standard derivation and mean of a set of values.
The first term ensures larger distinction between pixels of lit and shadow regions and the second term ensures smaller variation for pixels of the same lit or shadow regions.\newline
\textbf{2) Suppression of Texture} The noise due to image texture is reduced by applying a median filter with a $h_{2}$-by-$h_{2}$ neighborhood to $\mathcal{F}$.

\begin{figure}[htb]
  \centering
  \vspace{-5pt}
  \subfigure[RGB]{
    \includegraphics[width=0.45\linewidth]{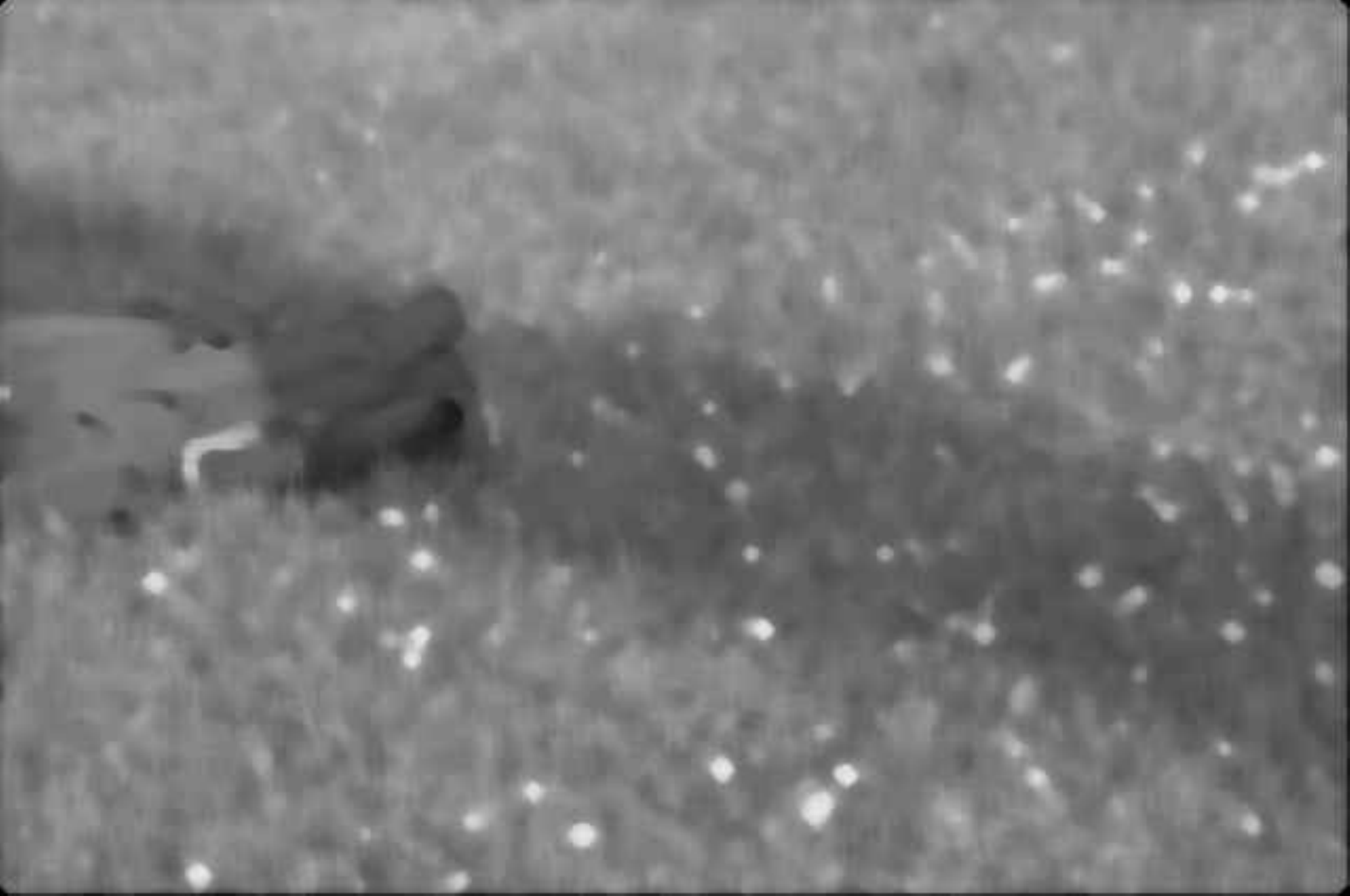}
  }
  \subfigure[YCrCb]{
    \includegraphics[width=0.45\linewidth]{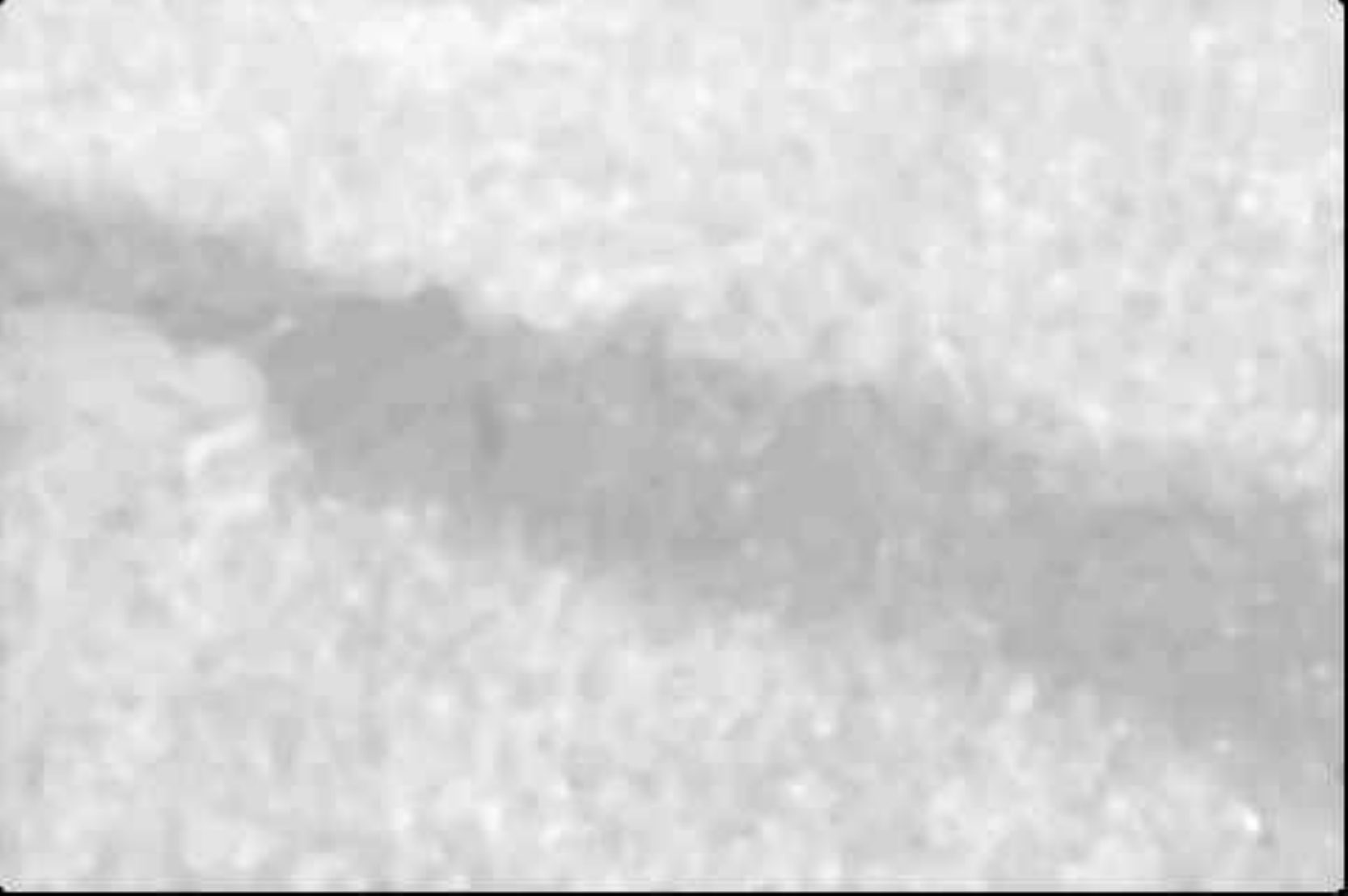}
  }
  \vspace{-5pt}
  \caption{Comparison of fusion image using different color spaces. The same optimization scheme is applied to the image in Fig.~\ref{fig:ppl}(a) but using different color spaces. The YCrCb fusion image presents more clean illumination information.}
  \label{fig:fusp}
\end{figure}

YCbCr color space offers perceptually meaningful information. Empirically, illumination information appears dominantly in one of its channels. The illumination information in RGB channels are usually affected by texture noise. An example of comparison between fusing channels using YCrCb color space and RGB color pace is shown in Fig.~\ref{fig:fusp}.
\subsection{Penumbra Unwrapping}
\label{sec:pu}
A shadow boundary generally has a noisy profile with a variable penumbra width. This can lead to inaccurate estimation of shadow scales and resulting artifacts. The penumbra is therefore unwrapped into a strip and its sampled columns of illumination change are aligned (\eg Fig.~\ref{fig:ppl}(f)). This improves the detection of outliers and allows linearization of processing along the penumbra -- leading to significant gains in efficiency and speed.

\setlength{\textfloatsep}{0pt}
\begin{algorithm2e}
 \SetAlgoLined
 \SetKwData{Left}{left}\SetKwData{This}{this}\SetKwData{Up}{up}
 \SetKwFunction{Union}{Union}\SetKwFunction{FindCompress}{FindCompress}
 \SetKwInOut{Input}{input}\SetKwInOut{Output}{output}

 \Input{boundary point $(x_{b},y_{b})$, fusion image $\mathcal{F}$}
 \Output{two ends ($\mathbf{p}_{s}$, $\mathbf{p}_{e}$) of a sampling line}
 $\widetilde{F} \gets \nabla \mathcal{F}$;
 $\mathbf{p}_{s} \gets (x_{b},y_{b})$;
 $\mathbf{p}_{e} \gets (x_{b},y_{b})$;
 $\mathcal{L} \gets  |\widetilde{F}(x_{b},y_{b})|$;
 $\Delta \mathbf{v} \gets  \widetilde{F}(x_{b},y_{b})/\mathcal{L}$\;
 \Repeat{$\mathbf{p}_{s}$ or $\mathbf{p}_{e}$ is not within the range of $\mathcal{F}$ \textbf{or} $h_{5}\mathcal{L}_{s}>\mathcal{L}$ or $h_{5}\mathcal{L}_{e}<\mathcal{L}$}{
  $\mathbf{v}_{s} \gets \widetilde{F}([\mathbf{p}_{s}])$;
  $\mathbf{v}_{e} \gets \widetilde{F}([\mathbf{p}_{e}])$\;
  $\mathcal{L}_{s} \gets \mathbf{v}_{s} \cdot \Delta \mathbf{v}$; $\mathcal{L}_{e} \gets \mathbf{v}_{e} \cdot \Delta \mathbf{v} $\;
   $\mathbf{p}_{s} \gets \mathbf{p}_{s} - \Delta \mathbf{v}$; $\mathbf{p}_{e} \gets \mathbf{p}_{e} + \Delta \mathbf{v}$\;
 }
 \caption{Penumbra Sample End Point Selection}
 \label{agl:sample_search}
\end{algorithm2e}
Similar to prior work~\cite{Arbel2011,khan2016automatic}, the intensity of sampling lines perpendicular to the shadow boundary (Fig.~\ref{fig:ppl}(d)) are sampled. The length of a sampling line is determined by locating suitable start and end points guided by the fusion image $\mathcal{F}$. A bi-directional search is initialized from each boundary point that extends the sampling line towards the lit area (end point) and the shadow area (start point) as described in Algorithm~\ref{agl:sample_search}. This extension is symmetric. The start and end points are initially set as the boundary point $(x_{b},y_{b})$ and the direction vector $\Delta \mathbf{v}$ as the normalized gradient vector of $(x_{b},y_{b})$. To get the position for a start point, $\Delta \mathbf{v}$ is iteratively subtracted from the start point until its projected gradient is small enough (vice versa for the end point). Absolute gradient magnitude is not used in this algorithm because the gradient magnitudes of the soft penumbra edges can be very insignificant. The length of a sampling line thus depends on the starting gradient strength at the middle of penumbra. 

Instead of processing unaligned and unselected samples individually~\cite{Arbel2011,khan2016automatic}, we transform these samples into unified columns of the initial penumbra strip to enable fast batch processing. And, only the good samples are kept for shadow scale estimation. To avoid outliers, \eg sampling lines at occlusion boundaries, invalid samples are filtered based on an assumption of similar shadow scales. A scale vector $Y_{c}=T_{l}-T_{s}$ is first computed where $T_{l}$ and $T_{s}$ are the average Log-domain RGB intensities of the lit and shadow halves of a sampling line. $Y_{c}$ is then converted to spherical coordinates and considered as feature vector $Y_{s}$. DBSCAN clustering~\cite{Ester96adensity-based} (radius: $h_{3}$) is applied to $Y_{s}$ for all samples, and samples that belong to the largest cluster are stored as valid ones with valid illumination. For finer scale estimation, valid clusters are further divided into a few sub-groups using mean-shift~\cite{comaniciu2002mean} (band width: $h_{4}$) and the samples of invalid sub-groups, whose total members are less than $10\%$ of the largest sub-group's, are discarded. Fig.~\ref{fig:ppl}(d) shows an example of the above outlier detection.

\begin{figure}[htb!]
   \centering
   \resizebox{\linewidth}{!}{\input{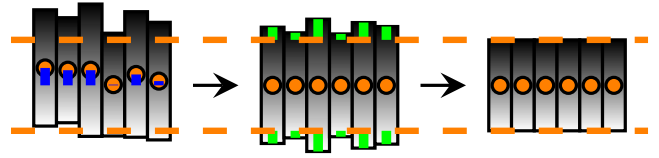}}
   \caption{Alignment of penumbra strip: The orange circles are the centers of columns in the penumbra strip. The orange dashed lines indicate the desired column length for a strip. The blue and green lines indicate the shifts required. In each iteration of the optimization, the alignment is in two steps: (left to middle) alignment of center; (middle to right) alignment of illumination change.}
   \label{fig:aligncol}
 \end{figure}

To achieve an efficient batch shadow scale estimation, we need to cancel out the affection of variable penumbra width. The lengths of samples are normalized by re-sizing all the samples to a unique length $n_{a}$ which is the maximum length of all valid samples. The normalized samples are then concatenated as columns to form the initial penumbra strip. Although our previous adaptive sampling already provides the intensity profiles with roughly aligned illumination changes, some minor errors may still exist. It is assumed that, after the previous intensity outlier filtering and the sample length normalization, the trends of intensity changes are similar and the dissimilarity only appears on the level of noise (\eg background texture) and some minor alignment errors. This is resolved by a fine-scale alignment by optimization. For each column (intensity profile), the alignment process vertically shifts the column's center and then stretches the column about its center by shifting its two ends. An illustration of this alignment is shown in Fig.~\ref{fig:aligncol}.
The parameters of this fine-scale alignment for each column are estimated by minimizing the following energy function $E_{a}$:
\begin{gather}
    L_{n} = \Gamma(A_{s},A_{k},L_{o}) \\
    E_{a} = \text{MSE}(L_{n} - L_{a})
\end{gather}
where $A_{s}$ and $A_{k}$ are the stretching shift and the center shift in the fine-scale alignment respectively, $L_{o}$ is the scales of original column, $L_{a}$ is the reference of alignment which is the average scale values of all valid columns (\ie column-wise mean of the penumbra unwrap), $L_{n}$ is the aligned unwrap, $\Gamma$ is a function that aligns $L_{o}$ according to the estimated alignment parameters, $\text{MSE}$ is a function that computes mean squared error. The minimization is solved using a sequential quadratic programming algorithm~\cite{nocedal2006numerical}.

\subsection{Relighting}
\label{sec:ess}
Using the aligned penumbra unwrap, a fast multi-scale shadow scale estimation is adopted for each shadow boundary. Compared with \cite{Liu2008,Su2010,Gong2013}, our shadow scale estimation is fast and adaptive, which neither requires computational-costly pixel-wise optimization nor assumes any constrained data fitting models of illumination change, \eg cubic curves or surface models. The non-linear image post-processing can significantly distort the original shadow scale change. Also, complex lighting conditions make the penumbra shadow scale change unpredictable. Instead, we only assume that illumination change is smooth and surface material change causes sharp intensity variation. Our recovery does not constrain the shape of the smooth shadow scale change of a sampling line. This means that our penumbra recovery is compatible with a wider range of shadow scale changes (\eg ours can remove unconventional colored shadows) as long as the shadow scale changes of neighboring sampling lines are not too dissimilar. Our previous sample length normalization and alignment make it possible to estimate the shadow scale change by a simple and adaptive horizontal smoothing.

\begin{figure}[htb!]
  \centering
  \includegraphics[width=\linewidth]{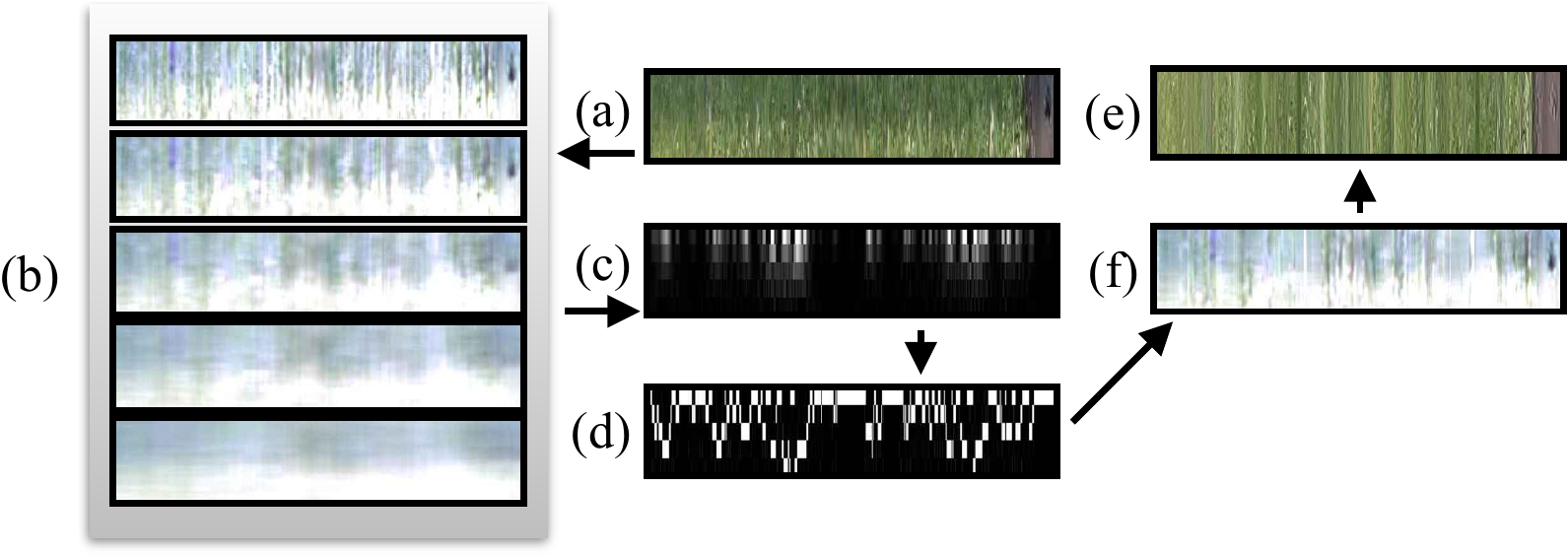}
  \caption{Pipeline of multi-scale shadow scale estimation. The aligned penumbra unwrap (a) is filtered using average kernels in exponentially increasing sizes to build a pyramid of shadow scales (b). The roughness of each column of each pyramid layer is measured and visualized in (c). Brighter colors indicate higher roughness. The horizontal and vertical dimensions in (c) refer to column index and layer index respectively. (d) is the visualized corresponding selections of layer index for each column (in white) after thresholding. (e) is the finale shadow scale composed by the shadow scales from the different layers of (b). (f) is the relit penumbra unwrap using (e).}
  \label{fig:mse}
\end{figure}

A pyramid (\eg Fig.~\ref{fig:mse}(b)) of horizontally filtered penumbra unwraps using 5 averaging kernels in different sizes are computed so that texture noise can be canceled. The sizes of averaging kernels are specified as 1-by-$2^{\tilde{n}}$ where $\tilde{n} \in \{2,3,4,5,6\}$. The filtered intensities of the pyramid are then converted to shadow scales. For each RGB channel layer of each pyramid layer, the estimated scales can be computed by dividing the intensities of each column by the intensity of the last element of each column (\ie lit end). The optimum shadow scales for each column are selected from different layers of the pyramid. Column intensity with higher localness (\ie filtered by a smaller kernel) and lower roughness are preferred. However, higher localness leads to higher roughness, so an optimum solution should balance these two properties. The roughness of intensity change $E_{s}(\tilde{c},\tilde{n})$ (visualized in Fig.~\ref{fig:mse}(c)) is measured as follows:
\begin{equation}
    E_{s}(\tilde{c},\tilde{n}) = \int { \left(\frac{\partial^{2} U(\tilde{r},\tilde{c},\tilde{n})}{\partial \tilde{r}^{2}}\right)^{2}} d\tilde{r}
\end{equation}
where $U$ is the penumbra unwrap, $\tilde{n}$ is the layer index of pyramid, $\tilde{c}$ and $\tilde{r}$ are the column and row coordinates of the penumbra unwrap respectively. The optimum scales for each column are selected using a threshold of roughness $T_{s}$ which is computed as the mean of all values in $E_{s}$. The column of one of the layers which has the lowest roughness above $T_{s}$ is selected (visualized in Fig.~\ref{fig:mse}(d)). A shadow scale image of the penumbra unwrap (\eg Fig.~\ref{fig:mse}(e)) can thus be formed by picking columns from different pyramid layers according to the selected layer index of each column. As the intensity samples, \ie columns, have previously been aligned during the alignment of the unwrap, the estimated scales of each sampling are mapped back by using an reverse operation of $\Gamma$ so that the estimated shadow scales are corresponding to the original unaligned intensities of the penumbra unwrap. The mapped-back shadow scales are then registered to the their 2D positions in the image that a sparse shadow scale field is formed (\eg Fig.~\ref{fig:ppl}(g)). To obtain a dense scale field (\eg Fig.~\ref{fig:ppl}(h)), we propagate the sparse scales in the penumbra region by smoothly interpolating and extrapolating the scales in other regions using image in-painting~\cite{Bertalmio2000}. The shadow-free image can be obtained by inverse scaling according to Eq.~\ref{equ:image_formation}.

\subsection{Color Correction}
\label{sec:gcc}
Images captured from popular imaging devices are often post-processed, \eg gamma correction and JPEG compression, such that the linearity of photon intensity is not maintained. When the degree of post-processing is high, visible artifacts, \eg differences in tone and contrast, may appear in shadow corrected areas as Eq.~\ref{equ:image_formation} does not hold. This is because these post-processing steps apply non-linear operations which break the linearity property that the intensity of a pixel is proportional to the amount of photons a sensor has received. A robust multi-scale color correction method is therefore proposed to address this issue.
The improvement will only be significant for images which are over post-processed. Previous work has proposed global adjustments to align the intensity characteristics of the umbra and lit area~\cite{Gong2013,Liu2008}. These assume that the surface around the penumbra has a similar texture and color but may lead to significant unnatural artifacts when they are dissimilar. To address this, we adopt an image detail alignment similar to Xiao \etal~\cite{xiao2013fast} that equates the spatially dependent variance of RGB intensities between the shadow and lit sides at different scales. Unlike \cite{xiao2013fast}, our method does not require a shadow removal for each filtered image level (which is computationally expensive). Instead, based on an initial shadow removal result, we only iteratively align the variances around shadow boundary on each scale.
\begin{figure*}[htb]
   \centering
   \includegraphics[width=\linewidth]{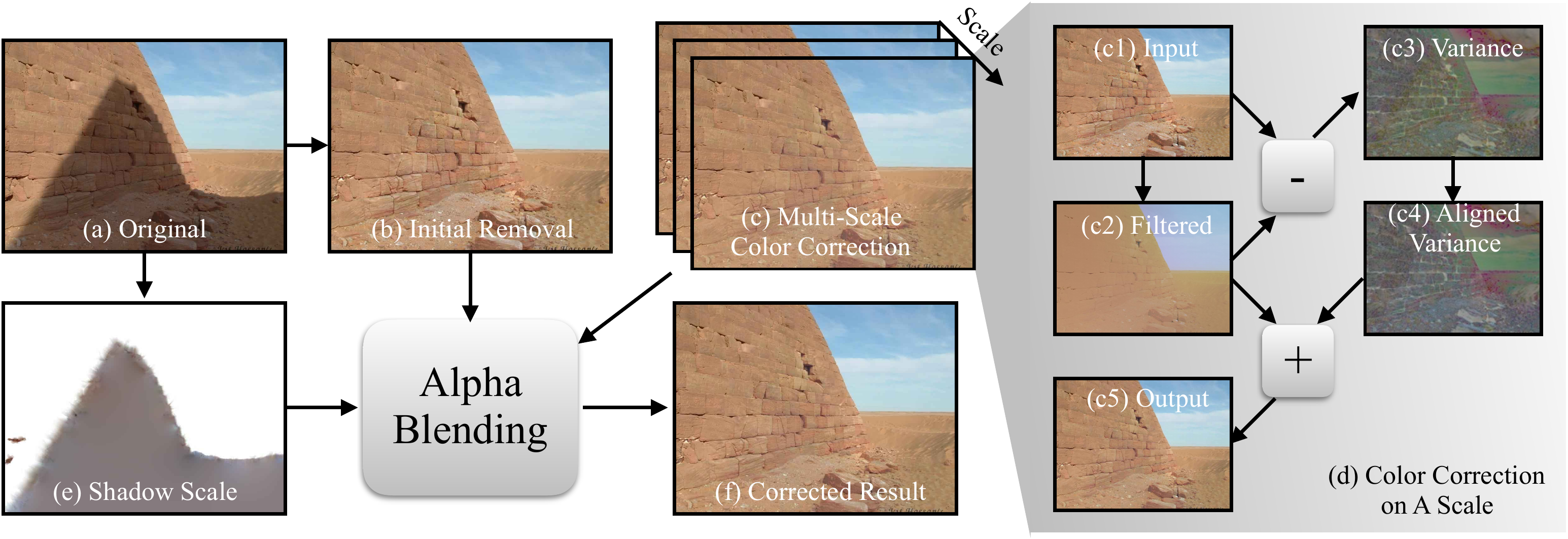}
   \caption{Multi-scale color correction pipeline. The inconsistency in the initial shadow-free image (b) is fixed in the final output (f). The multi-scale color correction aligns the color variance at different scales from coarse to fine. On each single scale, the initial input image (c1) exhibits inconsistency of local variance between lit and shadow areas. The higher-frequency variation (c3) of shadow and lit areas are aligned in (c4). The corrected output (c5) can be obtained by adding (c4) to (c2).}
   \label{fig:gcc}
\end{figure*}
It is assumed that the average intensity of both sides of the shadow are accurate and that artifacts are due to the differences in intensity variance. Statistics are collected from the lit side pixels $\mathbf{P}_{l}$ and the umbra side pixels $\mathbf{P}_{u}$ both near the penumbra as the reference and source of color correction respectively. The algorithm for alignment is described in Algorithm~\ref{agl:cc}. 
\begin{algorithm2e*}
 \SetAlgoLined
 \SetKwData{Left}{left}\SetKwData{This}{this}\SetKwData{Up}{up}
 \SetKwFunction{Union}{Union}\SetKwFunction{FindCompress}{FindCompress}
 \SetKwInOut{Input}{input}\SetKwInOut{Output}{output}

 \Input{shadow removed image $I^{r}$, reference lit pixels $\mathbf{P}_{l}$, source shadow pixels $\mathbf{P}_{u}$, all shadow pixels $\mathbf{P}_{s}$}
 \Output{color corrected image $I^{ra}_{c}$}
 
 $I^{ra} \gets I^{r}$\;
 \For(for each scale, \eg, Fig.~\ref{fig:gcc}){s = 1 \emph{\KwTo} 3}{
     $I^{l} \gets bfilter(I^{ra},\beta/2^{s+1},h_{6})$ \tcp{apply bilateral filtering (\eg Fig.~\ref{fig:gcc}(c2))}
   $I^{h} \gets I^{r} - I^{l}$ \tcp{get local intensity variation image (\eg Fig.~\ref{fig:gcc}(c3))}
   \For(for each RGB channels){c = 1 \emph{\KwTo} 3}{
       $r_{\sigma} \gets \varsigma(I^{h}_{c}(\mathbf{P}_{l}))/\varsigma(I^{h}_{c}(\mathbf{P}_{u}))$ \tcp{get overall ratio of intensity variation}
       $I^{re}_{c} \gets r_{\sigma}I^{h}(\mathbf{P}_{s}) $ \tcp{get aligned intensity variation image (\eg Fig.~\ref{fig:gcc}(c4))}
       $I^{ra}_{c} \gets I^{r}$ \tcp{copy intensities of lit pixels}
       $I^{ra}_{c}(\mathbf{P}_{s}) \gets I^{l}_{c}(\mathbf{P}_{s}) + I^{re}_{c} $ \tcp{add aligned intensity variation back (\eg Fig.~\ref{fig:gcc}(c5))}
   }
 }
 \caption{Multi-Scale Color Correction}
 \label{agl:cc}
\end{algorithm2e*}
where $s$ is a scale, $\beta$ is the maximum image dimensional of $I^{ra}$, $bfilter$ is an operation that bilaterally filters~\cite{ParisD09} the input image (first parameter) using a standard deviation of the space (second parameter) and a range Gaussian (third parameter), $I^{h}$ is an image of intensity variation, where $c$ is the channel index, $\varsigma$ is a function which computes the median absolute deviation.

Finally, to smooth the color correction result, alpha blending is applied in RGB color space according to the shadow scale as shown in Eq.~\ref{equ:bld}.
\begin{equation}
\label{equ:bld}
I^{f}_{c} = I^{r}_{c} \circ \dot{\mathcal{S}}_{c} + I^{ra}_{c} \circ (\mathds{1}-\dot{\mathcal{S}}_{c})
\end{equation}
where $c$ is the channel index, $\dot{\mathcal{S}}_{c}$ is the normalized scale field of $\mathcal{S}$, $I^{f}_{c}$ is the final shadow-free image. An illustration of the intermediate steps of color correction is shown in Fig~\ref{fig:gcc}.

\subsection{Parameter Learning}
\label{sec:para}
Our shadow removal approach includes the use of some standard algorithms that require the specification of parameters. To determine an appropriate set of parameters, we apply an optimization to learn these parameters from a subset of our ground truth data set. These parameters contain both integers and real numbers. It is therefore not possible to apply a gradient-based optimization method that requires the objective and constraint functions to be both continuous and have continuous first derivatives. To determine these parameters, we apply a mixed-integer genetic optimization method~\cite{deep2009real}. Let us define $H = h_{1},h_{2},\dots,h_{6}$ as a vector of the shadow removal parameters denoted throughout the paper. Our objective function $E_{p}$ minimizes the sum of all error measurements as the follows:
\begin{equation}
    E_{p}(H) = \sum_{k} e_{k}w_{k}
\end{equation}
where $e_{k}$ is the $k^{\text{\tiny th}}$ error measurement, and $w_{k}$ is the weight for $e_{k}$. We assume that the weights for all error measurements are the same (\ie equally important), \eg, $w_{k} = 1$. These error measurements are later introduced in \S\ref{sec:qeval} (in Table~\ref{tb:res}) and only all-pixel-error is used in our learning. Table~\ref{tbl:para} shows the details of these parameters and their optimization configuration. In our experiment, 5 test cases are randomly selected for computing each error measurement.
\begin{table}[htb!]
\caption{Parameter learning specification for the optimization.}
\label{tbl:para}
\input{para.clo}
\end{table}

The optimum parameters for all error measurements are learned as $\begin{bmatrix} 14&10&0.1124&0.0333&8.5195&0.2228\end{bmatrix}$. Our evaluation results in \S\ref{sec:ev} are reported using these learned optimum parameters. Our experiment demonstrates that it is possible to learn an optimum set of shadow removal parameters from a ground truth data set. However, the learned parameters are also dependent on the amount and quality of training data. We further discover the choice of parameters for different types of shadows. An individual parameter learning process is performed for each shadow category measurement, \ie, $E_{p}(H) = e_{k}$ where $k$ indicates the $k^{\text{\tiny th}}$ error measurement. Table~\ref{tbl:para_i} shows our parameter learning results for the individual shadow categories (error measurement). 
\begin{table*}[htb]
\small
\caption{Parameter learning results for individual error measurement.}
\label{tbl:para_i}
\input{para_i.clo}
\end{table*}
These individual parameter learning results show that the optimum parameters can vary depending on type of shadow. It is therefore practical to provide some predefined parameter sets for different shadow removal tasks or a single parameter set that balances the shadow removal performance for all types of shadows.

\section{Evaluation}
\label{sec:ev}
In this section, we first describe experiments that highlight our algorithms behavior given variable user inputs. The quality of our new ground truth versus existing state-of-the-art ground truth is then quantitatively evaluated. Finally, our algorithm is evaluated versus other state-of-the-art shadow removal methods based on our new dataset.

\subsection{Performance Stability Given Different User Inputs}
\label{sec:variable_input}
\begin{figure}[htb!]
\centering
\input{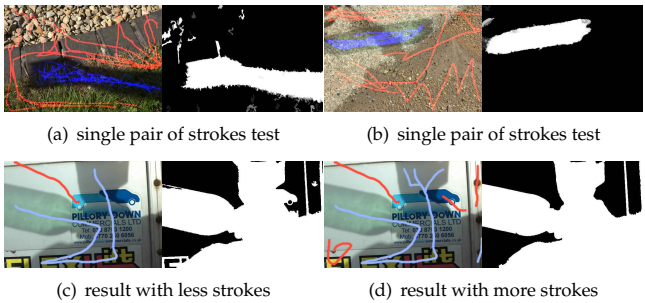}
\caption{Variable input behaviors: The top row shows two examples using single pairs of strokes. 10 examples of single strokes placed in different locations are supplied as input (red for lit and blue for shadow). The 2 gray-level images show the visualized probability of each pixel being marked in these 10 independent tests. Fewer gray pixels indicate higher stability, \ie the image should only show black (0\% probability) and white (100\% probability) pixels when it is absolutely stable. The bottom row shows examples highlighting how additional strokes can improve the detection result (binary mask).}
\label{fig:input1}
\end{figure}
Given user-supplied single pairs of strokes of lit and shadow pixel samples, our shadow detection generates stable results in different conditions (\eg Fig.~\ref{fig:dti1} and Fig.~\ref{fig:dti2}). In some cases, \eg where the surface color is very shadow-like, the detection results can be improved by supplying more than one pair of strokes (\eg Fig.~\ref{fig:dti31} and Fig.~\ref{fig:dti32}).

\begin{figure}[htb]
\centering
\input{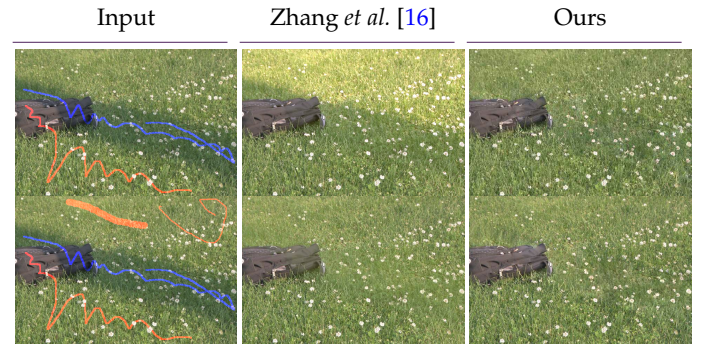}
\caption{Rougher stroke requirement: To generate a reasonable shadow removal result, our method requires less input strokes (in the same format) compared with \cite{zhang2015shadow}.}
\label{fig:input2}
\end{figure}

While our user input format is the same as the input format adopted in \cite{zhang2015shadow,xiao2013fast}, our method is found more robust to rougher (or fewer) scribbles since ours does not require an image matting process that is affected by sampled pixel location. Fig.~\ref{fig:input2} shows an example.

\subsection{Evaluation of Ground Truth Quality}
Ideal pairs of ground truth images should have a minimum intensity difference in the common lit area -- which will also indicate whether registration is poor (due to camera shake or scene movement -- which should be rejected). This is utilized to assess the quality of ground truth candidates. The error image $\Delta I = I_{s}-I_{g}$ and the ratio image $I_{r}=\Phi(I_{s})\oslash\Phi(I_{g})$ are first computed, where $I_{s}$ and $I_{g}$ are the original shadow image and its shadow-free ground truth image (which differs from the processed shadow-free outputs $I^{f}$ in Eq.~\ref{equ:bld}) respectively, $\oslash$ is element-wise division and $\Phi$ is a function that converts RGB image to gray-scale image. The set of pixels $P_{r}$ of $I_{r}$ that satisfies $I_{r}(P_{r})\geq 1$ are regarded as lit pixels. Due to some unavoidable minor global illumination changes and the inaccuracy in camera exposure control, the lit intensities in the shadow image can be higher than those in the shadow-free ground truth image. $I_{r}$ can therefore be greater than $1$. The ground truth error $Q_{d}$ is computed as follows:
\begin{equation}
\label{equ:gtsc}
Q_{d} = \mu(\abs{\Delta I(P_{r})}) + \sigma(\Delta I(P_{r}))
\end{equation}
Ground truth pairs in our data set with $Q_{d}>0.05$ are removed. Using this measure, our initial data capture of 195 test cases results in 186 test cases with stable illumination changes between the shadow and ground truth images. Comparing to the quality of other ground truth data sets, \cite{Guo2012} (state of the art) results in mean error of $0.18$ (leaving 28 out of 79 test cases) while ours is $0.02$.

\subsection{Quantitative evaluation of shadow removal}
\label{sec:qeval}
In previous work~\cite{ShorL2008,Guo2012}, the quality of shadow removal is measured by directly using the per-pixel error between the shadow removal result and shadow-free ground truth. However, a shadow in a smaller size or a lighter shadow can result in a smaller initial error between the original shadow image and its shadow-free ground truth. It is thus unfair to judge that a method is better only because the error between the shadow-removed image and its shadow-free ground truth is smaller. In our work, we cancel out the affects of the size and darkness of the shadow. We therefore compute the error ratio $\mathbf{E}_{r}$ as our quality measurement:
\begin{equation}
\mathbf{E}_{r} = \mathbf{E}_{n}/\mathbf{E}_{o}
\label{eq:error}
\end{equation}
where $\mathbf{E}_{n}$ is the error between the ground truth (no shadow) and shadow removal result, and $\mathbf{E}_{o}$ is the error between the ground truth (no shadow) and the original shadow image. This normalized measure better reflects removal improvements towards the ground truth independent of original shadow intensity and size. We assess $\mathbf{E}_{n}$ and $\mathbf{E}_{o}$ using Root-Mean-Square-Error (RMSE) of RGB intensity.
To test robustness, the standard deviation for each measurement is also computed. Unlike previous un-categorized test~\cite{ShorL2008,Guo2012}, our removal test is based on our data set of 186 cases, which contains challenging soft, broken and color shadows and shadows cast on strong textured surfaces as well as simpler shadows, plus 28 remaining cases from \cite{Guo2012} -- resulting in 214 test cases in total. Each case is rated according to 4 attributes, which are \emph{texture}, \emph{brokenness}, \emph{colorfulness} and \emph{softness}, in 3 perceptual degrees from weak to strong which were aggregated by 5 users.
\begin{table*}[htb]
\centering
\small
\caption{Shadow removal errors for test cases according to four attributes. The left and right sides of the table show the error scores where all pixels in the image are used, and just shadow area pixels respectively. For each score of each attribute, the images with other predominant attributes (strong) are not used. Hence, test cases have a strong single bias towards one of the attributes. ``Other'' refers to a set of shadow cases showing no markedly predominant attributes. ``Mean'' refers to the average score for each category. Standard derivations are shown in brackets. In our ordering, the average error is compared before comparing the standard derivation. Method \cite{Guo2012} is trained using a large shadow detection data set from~\cite{Zhu2010}. The user input for Method \cite{zhang2015shadow} is a combination of the simple input for our method and some additional strokes for accommodating the sensitive shadow detection of \cite{zhang2015shadow}. The best scores are made bold.}
\label{tb:res}
\input{num_w.clo}
\end{table*}
In Table~\ref{tb:res} (left side), the combined shadow removal error results from both automatic and semi-automatic shadow removal algorithms (all 214 cases) are shown.
We separate the error measurement for all pixels and only shadow pixels. In our experiments, our method shows significant leading performance across all metrics.
According to Eq.~\ref{eq:error}, $E_{o}$ for all pixels is lower than $E_{o}$ for the shadow pixels only because the intensity errors of the lit pixels are close to $0$ and $E_{o}$ measures the RMSE. In addition, the new RMSE $E_{n}$ for both all pixels and shadow pixels only are very close after shadow removal, the error ratio $E_{r}$ for the shadow pixels only is therefore generally lower.


To encourage open comparison in the community, we provide an online benchmark\footnote{\url{http://cs.bath.ac.uk/\%7Ehg299/shadow_eval/}} for quantitative evaluation of shadow removal.

\subsection{Analysis of Shadow Categories and Attributes}
\begin{figure}[htb!]
  \centering
  \input{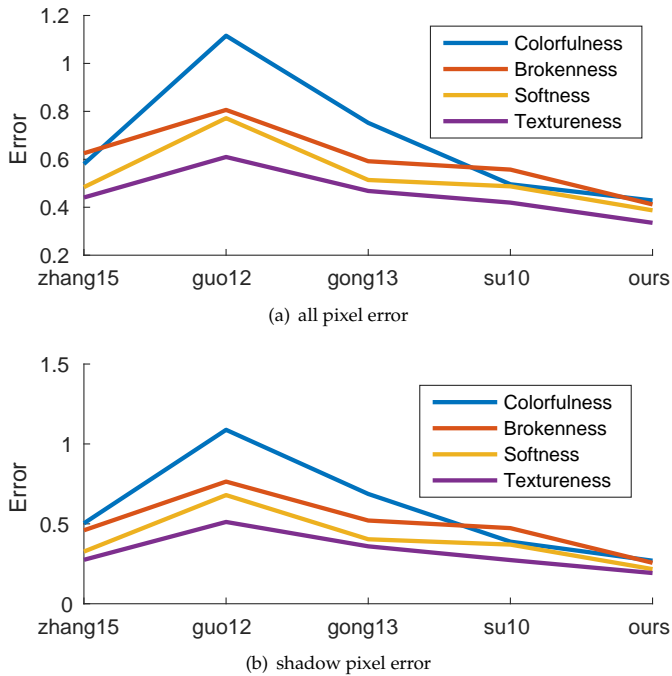}
  \caption{Parallel coordinate charts of the quantitative results in Table~\ref{tb:res}. The ticks zhang15, guo12, gong13, and su10 refer to \cite{zhang2015shadow}, \cite{Guo2012}, \cite{Gong2013}, and \cite{Su2010} accordingly. The scores presented here are the average scores of all three degrees for each attribute.}
  \label{fig:aa}
\end{figure}
To investigate the affects of different shadow categories and attributes, the quantitative result in Table~\ref{tb:res} is summarized by visualizing the result using the parallel coordinate charts in Fig.~\ref{fig:aa}. Such a visualization is insightful as strong performance of one method could direct practitioners to favor one algorithm over another in some problem cases. Overall, colored shadows are shown to be significantly the most difficult shadows to remove and shadows cast on high texture the easiest challenge. Broken shadows are slightly more difficult to process than soft shadows, although both of them are in the range of medium difficulty. \cite{Guo2012} and \cite{Gong2013} show relatively significant disadvantages in processing color shadows, while \cite{Su2010} demonstrates obvious difficulty in processing broken shadows. The trend of the other methods and attributes are otherwise similar. In our tests, our method overall demonstrates the best performance for all types of shadows analyzed.

\subsection{Visual Comparison}
\label{sec:eva-vc}
\begin{figure*}[htb!]
  \centering
  \input{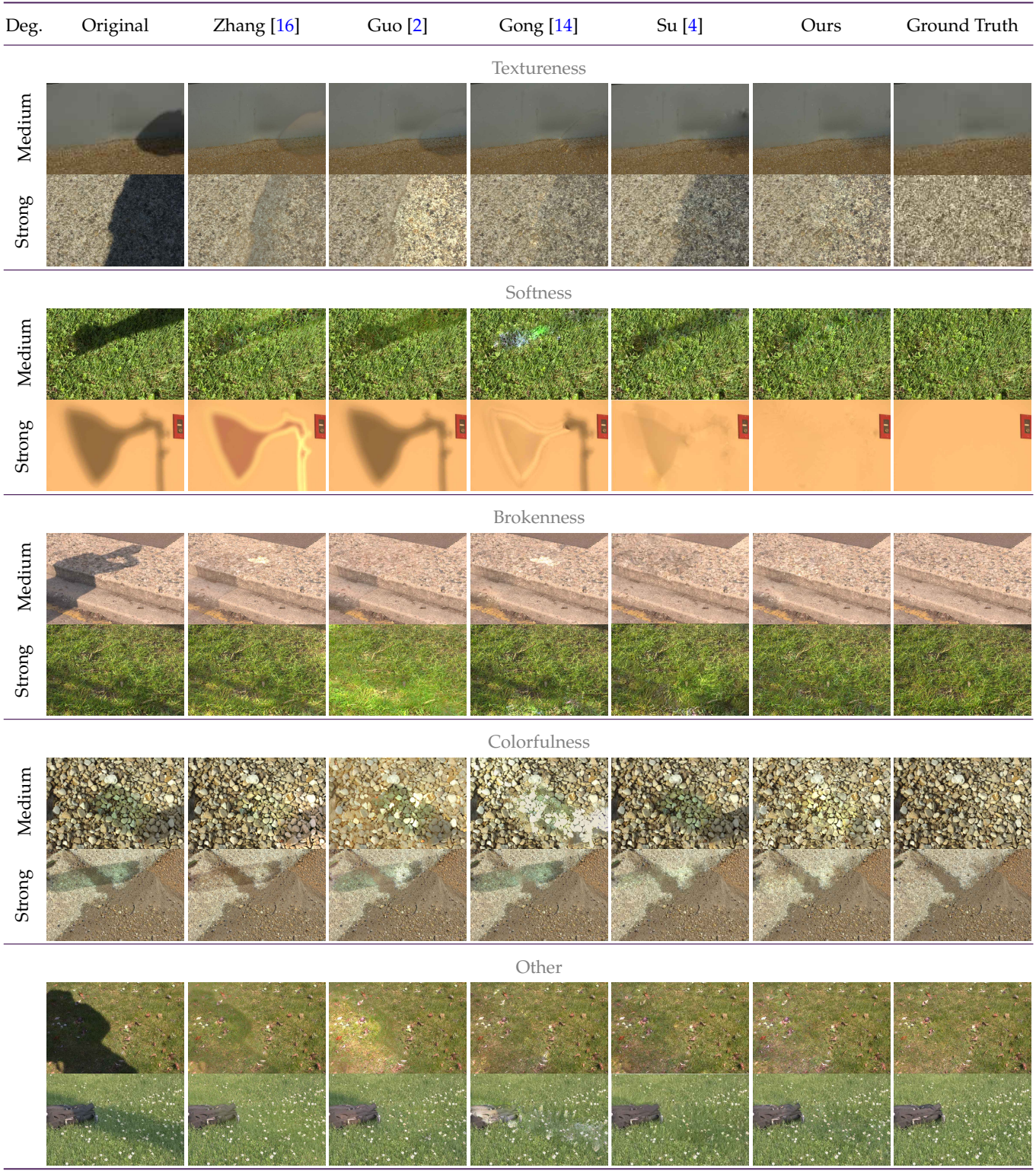}
  \caption{Comparisons using images from our data set. The table shows our results given test cases with stronger degrees of the corresponding attribute except for ''Other'', which refers to cases where there is no predominantly strong attribute. Degree weak is not shown as weak means almost no attribute feature.
	}
  \label{fig:datashow}
\end{figure*}

Fig.~\ref{fig:datashow} shows some typical visual results of shadow removal on various scenarios from our data set.

\subsection{Efficiency Comparison}
Table~\ref{tb:res_time} shows the required time for processing 0.3 mega-pixel color images shown in Table~\ref{fig:datashow} on a 3.1GHz machine. Our MATLAB implementation generally requires less system processing time than the other two MATLAB implementations of~\cite{Guo2012,Gong2013}, one MATLAB+C implementation~\cite{zhang2015shadow} and one C++ implementation~\cite{Su2010}. Compared with the other user-assisted methods~\cite{Gong2013,Su2010}, our method also generally requires less time for user-interaction. The slower performance of \cite{zhang2015shadow} is majorly caused by its slow image matting pre-processing step.
\begin{table}[htb!]
\small
\centering
\caption{Time comparison of shadow removal: $T_{u}$, $T_{s}$ refer to time (in seconds) for user interaction and time for system processing respectively.}
\label{tb:res_time}
\input{num_time.clo}
\end{table}

\subsection{Limitation And Future Work}
\begin{figure*}[htb!]
  \centering
  \small
  \input{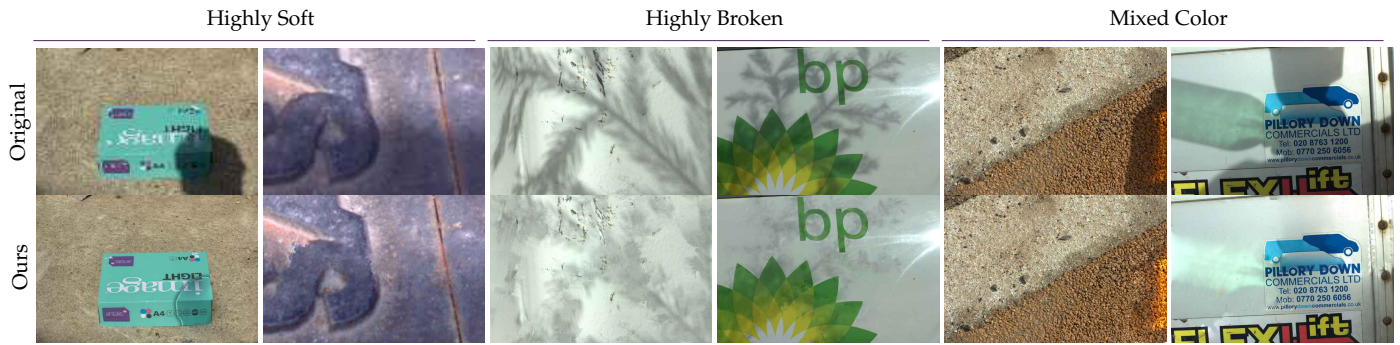}
  \caption{Failure cases -- where all the leading methods tested result in visible image artifacts or fail to remove the shadow. These images are also included in our shadow removal dataset published online.}
  \label{fig:datashow_f}
\end{figure*}

As is the case with all current shadow removal methods, our algorithm has most difficulty in extreme cases, \eg Fig.~\ref{fig:datashow_f}, where shadows are highly soft, broken, colorful (mixed by at least two different colors). Although the shadow effects can be significantly reduced by our method, the artifacts are still noticeable. A shadow image may also contain a mixture of more than one strong shadow attributes (\eg the last column of Fig.~\ref{fig:datashow_f}). We have challenged the multi-category shadow removal for the first time in our community (with a leading performance), but have not resolved the extreme cases. This highlights a direction for future work.

Our method requires users to supply reasonable inputs. We have not considered its tolerance for very careless user inputs (\eg mistakenly marking many shadow pixels as lit samples). Besides, insufficient user inputs may result in a poor shadow detection. Since a sufficiently trained shadow classifier may be robust to this issue, another future work could be improving shadow detection by combining user inputs with the shadow masks generated from an automatic shadow classifier.

\section{Conclusions}
We have presented an interactive method for fast shadow removal together with a state-of-the-art ground truth. Our method balances the complexity of user input with robust shadow removal performance. Our quantitatively-verified ground truth data set overcomes issues of mismatched illumination and registration in existing data sets. We have evaluated our method against several state-of-the-art methods using a thorough quantitative test and shown leading state of the art performance. Besides the opportunities for improving shadow removal quality for the categorized shadows in our dataset, the detection and removal for highly-complicated shadows, such as overlapping shadows caused multiple light sources with different light colors, and shadows caused by transparent objects with complicated inner structure and color, is still an open problem for the community.


\bigskip
\bibliography{biblo.bib}   

\begin{thebibliography}{10}
\newcommand{\enquote}[1]{``#1''}

\bibitem{Arbel2011}
E.~Arbel and H.~Hel-Or, \enquote{Shadow removal using intensity surfaces and
  texture anchor points,} Transaction on Pattern Analysis and Machine
  Intelligence \textbf{33}, 1202--1216 (2011).

\bibitem{Guo2012}
R.~Guo, Q.~Dai, and D.~Hoiem, \enquote{Paired regions for shadow detection and
  removal,} Transaction on Pattern Analysis and Machine Intelligence
  \textbf{PP}, 1--1 (2012).

\bibitem{khan2016automatic}
S.~H. Khan, M.~Bennamoun, F.~Sohel, and R.~Togneri, \enquote{Automatic shadow
  detection and removal from a single image,} transactions on pattern analysis
  and machine intelligence \textbf{38}, 431--446 (2016).

\bibitem{Su2010}
Y.-F. Su and H.~H. Chen, \enquote{A three-stage approach to shadow field
  estimation from partial boundary information,} Transaction on Image
  Processing \textbf{19}, 2749--2760 (2010).

\bibitem{Liu2008}
F.~Liu and M.~Gleicher, \enquote{Texture-consistent shadow removal,} in
  \enquote{European Conference on Computer Vision,}  (Springer, 2008), pp.
  437--450.

\bibitem{ShorL2008}
Y.~Shor and D.~Lischinski, \enquote{The shadow meets the mask: Pyramid-based
  shadow removal,} Computer Graphics Forum \textbf{27}, 577--586 (2008).

\bibitem{barrow1978recovering}
H.~Barrow and J.~Tenenbaum, \enquote{Recovering intrinsic scene
  characteristics,} Computer Vision System, A Hanson \& E. Riseman (Eds.) pp.
  3--26 (1978).

\bibitem{Finlayson2009}
G.~D. Finlayson, M.~S. Drew, and C.~Lu, \enquote{Entropy minimization for
  shadow removal,} International Journal of Computer Vision \textbf{85}, 35--57
  (2009).

\bibitem{fredembach2006simple}
C.~Fredembach and G.~Finlayson, \enquote{Simple shadow removal,} in
  \enquote{International Conference on Pattern Recognition,} , vol.~1 (IEEE,
  2006), vol.~1, pp. 832--835.

\bibitem{Yang2012}
Q.~Yang, K.-H. Tan, and N.~Ahuja, \enquote{Shadow removal using bilateral
  filtering,} Transaction on Image Processing \textbf{21}, 4361--4368 (2012).

\bibitem{Drew2006}
M.~Drew, C.~Lu, and G.~Finlayson, \enquote{Removing shadows using flash/noflash
  image edges,} in \enquote{International Conference on Multimedia and Expo,}
  (IEEE, 2006), pp. 257--260.

\bibitem{Salamati2011}
N.~Salamati, A.~Germain, and S.~Susstrunk, \enquote{Removing shadows from
  images using color and near-infrared,} in \enquote{International Conference
  on Image Processing,}  (IEEE, 2011), pp. 1713--1716.

\bibitem{Wu2007}
T.-P. Wu, C.-K. Tang, M.~S. Brown, and H.-Y. Shum, \enquote{Natural shadow
  matting,} ACM Transactions on Graphics \textbf{26}, 8 (2007).

\bibitem{Gong2013}
H.~Gong, D.~Cosker, C.~Li, and M.~Brown, \enquote{User-aided single image
  shadow removal,} in \enquote{International Conference on Multimedia and
  Expo,}  (IEEE, 2013), pp. 1--6.

\bibitem{xiao2013fast}
C.~Xiao, R.~She, D.~Xiao, and K.-L. Ma, \enquote{Fast shadow removal using
  adaptive multi-scale illumination transfer,} in \enquote{Computer Graphics
  Forum,} , vol.~32 (Wiley Online Library, 2013), vol.~32, pp. 207--218.

\bibitem{zhang2015shadow}
L.~Zhang, Q.~Zhang, and C.~Xiao, \enquote{Shadow remover: Image shadow removal
  based on illumination recovering optimization,} Transactions on Image
  Processing \textbf{24}, 4623--4636 (2015).

\bibitem{fredembach2005hamiltonian}
C.~Fredembach and G.~Finlayson, \enquote{Hamiltonian path-based shadow
  removal,} in \enquote{British Machine Vision Conference,} , vol.~2 (BMVA,
  2005), vol.~2, pp. 502--511.

\bibitem{Huang2009}
J.-B. Huang and C.-S. Chen, \enquote{Moving cast shadow detection using
  physics-based features,} in \enquote{Conference on Computer Vision and
  Pattern Analysis,}  (IEEE, 2009), pp. 2310--2317.

\bibitem{Zhu2010}
J.~Zhu, K.~G.~G. Samuel, S.~Masood, and M.~F. Tappen, \enquote{Learning to
  recognize shadows in monochromatic natural images,} in \enquote{Conference on
  Computer Vision and Pattern Analysis,}  (IEEE, 2010).

\bibitem{Lalonde2010}
J.-F. Lalonde, A.~A. Efros, and S.~G. Narasimhan, \enquote{Detecting ground
  shadows in outdoor consumer photographs,} in \enquote{European Conference on
  Computer Vision,}  (Springer, 2010), pp. 322--335.

\bibitem{vicente2013single}
T.~F.~Y. Vicente, C.-P. Yu, and D.~Samaras, \enquote{Single image shadow
  detection using multiple cues in a supermodular mrf,} in \enquote{British
  Machine Vision Conference,}  (BMVA, 2013).

\bibitem{huang2011characterizes}
X.~Huang, G.~Hua, J.~Tumblin, and L.~Williams, \enquote{What characterizes a
  shadow boundary under the sun and sky?} in \enquote{International Conference
  on Computer Vision,}  (IEEE, 2011), pp. 898--905.

\bibitem{shen2015shadow}
L.~Shen, T.~Wee~Chua, and K.~Leman, \enquote{Shadow optimization from
  structured deep edge detection,} in \enquote{Conference on Computer Vision
  and Pattern Recognition,}  (IEEE, 2015), pp. 2067--2074.

\bibitem{Finlayson2007}
G.~Finlayson, C.~Fredembach, and M.~Drew, \enquote{Detecting illumination in
  images,} in \enquote{International Conference on Computer Vision,}  (IEEE,
  2007), pp. 1--8.

\bibitem{mitchell1997machine}
T.~M. Mitchell, \enquote{Machine learning. wcb,}  (1997).

\bibitem{Finlayson2002}
G.~D. Finlayson, S.~D. Hordley, and M.~S. Drew, \enquote{Removing shadows from
  images using retinex,} in \enquote{Color Imaging Conference,}  (2002), pp.
  73--79.

\bibitem{finlayson2006removal}
G.~D. Finlayson, S.~D. Hordley, C.~Lu, and M.~S. Drew, \enquote{On the removal
  of shadows from images,} Transactions on Pattern Analysis and Machine
  Intelligence \textbf{28}, 59--68 (2006).

\bibitem{finlayson2002removing}
G.~D. Finlayson, S.~D. Hordley, and M.~S. Drew, \enquote{Removing shadows from
  images,} in \enquote{European Conference on Computer Vision,}  (Springer,
  2002), pp. 823--836.

\bibitem{Ester96adensity-based}
M.~Ester, H.-P. Kriegel, J.~Sander, and X.~Xu, \enquote{A density-based
  algorithm for discovering clusters in large spatial databases with noise.} in
  \enquote{KDD,} , vol.~96 (ACM, 1996), vol.~96, pp. 226--231.

\bibitem{comaniciu2002mean}
D.~Comaniciu and P.~Meer, \enquote{Mean shift: A robust approach toward feature
  space analysis,} Transaction on Pattern Analysis and Machine Intelligence
  \textbf{24}, 603--619 (2002).

\bibitem{nocedal2006numerical}
J.~Nocedal and S.~Wright, \enquote{Numerical optimization,}  (Springer, 2006),
  Springer series in operations research, chap.~18, 2nd ed.

\bibitem{Bertalmio2000}
M.~Bertalmio, G.~Sapiro, V.~Caselles, and C.~Ballester, \enquote{Image
  inpainting,}  (ACM, 2000), pp. 417--424.

\bibitem{ParisD09}
S.~Paris and F.~Durand, \enquote{A fast approximation of the bilateral filter
  using a signal processing approach,} International Journal of Computer Vision
  \textbf{81}, 24--52 (2009).

\bibitem{deep2009real}
K.~Deep, K.~P. Singh, M.~Kansal, and C.~Mohan, \enquote{A real coded genetic
  algorithm for solving integer and mixed integer optimization problems,}
  Applied Mathematics and Computation \textbf{212}, 505--518 (2009).

\end{thebibliography}

\ifthenelse{\equal{\journalref}{ol}}{%
\clearpage
\bibliographyfullrefs{biblo.bib}
}{}

\end{document}